\documentclass[lettersize,journal]{IEEEtran}

\usepackage{amsmath,amsfonts,amsthm}

\usepackage{graphicx}
\usepackage{array,booktabs,multirow}
\usepackage{xcolor}
\usepackage{stfloats}

\usepackage{algorithm}
\usepackage{algorithmic}

\usepackage{cite}
\usepackage{url}
\usepackage{hyperref}
\usepackage{orcidlink}

\newcommand{\C}{\mathbb{C}}
\newcommand{\R}{\mathbb{R}}
\newcommand{\F}{\mathcal{F}}
\newcommand{\B}{\mathcal{B}}
\newcommand{\DC}{\operatorname{DC}}
\newcommand{\tr}{\operatorname{tr}}
\newcommand{\norm}[1]{\left\lVert #1\right\rVert}
\newcommand{\best}[1]{\textcolor{teal}{#1}}
\newtheorem{proposition}{Proposition}

\begin{document}

\title{STAIRS: Staging the Inverse Problem via Admissible-Support Scheduling for Unrolled MRI Reconstruction}

\author{Xiangjian~Hou\,\orcidlink{0000-0001-6092-3913}, Chang~Ni\,\orcidlink{0009-0009-0636-2735}, Chao~Qin\,\orcidlink{0009-0000-5014-4510}, Zhiwei~Tan\,\orcidlink{0000-0002-7964-0283}, Xin~Wang\,\orcidlink{0000-0001-9546-6038},\\
        Xinrui~Ye\,\orcidlink{0009-0005-1448-8195}, Rong-Rong~Chen\,\orcidlink{0009-0003-1462-5095}, Chun~Yuan\,\orcidlink{0000-0002-6154-5106}, and~Xiaodong~Ma\,\orcidlink{0000-0001-7158-8073}%
\thanks{This work was supported in part by
VA Merit Grant under I01CX003005.
\textit{(Corresponding author: Xiangjian Hou.)}}%
\thanks{Xiangjian Hou is with the Department of Radiology and Imaging
Sciences and the Department of Electrical and Computer Engineering,
University of Utah, Salt Lake City, UT 84112, USA
(e-mail: xiangjian.hou@utah.edu).}%
\thanks{Chang Ni is with the Department of Radiology and Imaging
Sciences and the Department of Biomedical Engineering, University of
Utah, Salt Lake City, UT 84112, USA (e-mail: chang.ni@utah.edu).}%
\thanks{Chao Qin is with the Department of Computer Vision,
Mohamed bin Zayed University of Artificial Intelligence, Abu Dhabi,
United Arab Emirates (e-mail: chao.qin@mbzuai.ac.ae).}%
\thanks{Zhiwei Tan and Xin Wang are with the Department of Electrical
and Computer Engineering, University of Washington, Seattle, WA 98195,
USA (e-mail: zhiwei24@uw.edu; xwang99@uw.edu).}%
\thanks{Xinrui Ye and Rong-Rong Chen are with the Department of
Electrical and Computer Engineering, University of Utah, Salt Lake
City, UT 84112, USA (e-mail: u1531017@utah.edu;
rchen@ece.utah.edu).}%
\thanks{Chun Yuan and Xiaodong Ma are with the Department of Radiology
and Imaging Sciences, University of Utah, Salt Lake City, UT 84112,
USA (e-mail: chun.yuan@hsc.utah.edu; xiaodong.ma@hsc.utah.edu).
Chun Yuan is also with the Department of Radiology, University of
Washington, Seattle, WA 98195, USA (e-mail: cyuan@uw.edu).}%
}

\markboth{Journal of \LaTeX\ Class Files,~Vol.~14, No.~8, August~2021}%
{Shell \MakeLowercase{\textit{et al.}}: A Sample Article Using IEEEtran.cls for IEEE Journals}


\maketitle

\begin{abstract}
Accelerated magnetic resonance imaging (MRI) forms images from incomplete Fourier measurements, shifting acquisition burden to computational reconstruction. Deep unrolled networks interleave learned regularization and data consistency, yet typically estimate the entire unmeasured spectrum at every cascade. We propose STAIRS (Staging the Inverse Problem via Admissible-Support Scheduling), which projects intermediate reconstructions onto nested Fourier supports that expand to the full spectrum and conditions each update on its support transition. Support expansion is scheduled in a normalized spectral coordinate built from conditional increments of a regularized log-determinant score of training covariance. These increments account for dependencies between newly and previously admitted frequencies. Dyadic targets allocate larger coordinate increments to earlier cascades and are mapped to fixed supports before training, yielding coarse-to-fine reconstruction while retaining the acquisition mask and data-consistency operator. On fastMRI and Stanford2D, STAIRS improves matched Transformer (HUMUSNet) and convolutional (NAFNet) backbones in single- and multi-coil reconstruction, with gains extending to radiologist-annotated pathology regions. STAIRS further outperforms predictor-based progression while reducing parameter overhead and inference latency. Stage-wise analysis shows repeated error correction by projection and lower intermediate reconstruction errors than the backbones alone throughout support expansion. Ablations favor dyadic allocation over uniform expansion for square and line-band supports and over manual front-loading for square windows. Support projection alone improves reconstruction, and conditioning on support transitions strengthens the gain. Cross-dataset evaluation shows improvements for both backbones in both directions. Support allocation thus complements regularizer design in unrolled MRI reconstruction.
\end{abstract}

\begin{IEEEkeywords}
Accelerated MRI, deep learning, image reconstruction, inverse problems, unrolled networks.
\end{IEEEkeywords}

\section{Introduction}
\label{sec:introduction}

\IEEEPARstart{M}{agnetic} resonance imaging (MRI) offers rich soft-tissue
contrast without ionizing radiation, but the spatial encoding required
to form an image is inherently time consuming. Accelerated MRI shortens
the examination by sampling only part of $k$-space. Parallel imaging
exploits the spatial diversity of receiver
coils~\cite{pruessmann1999sense,griswold2002generalized}, whereas
compressed sensing exploits sparsity in a suitable transform domain to
recover images from undersampled $k$-space below the Nyquist
rate~\cite{lustig2007sparse}. Both approaches shift part of the
acquisition burden toward computation, formulating image reconstruction
as an inverse problem constrained by known encoding physics and
undersampled data.

Deep unrolled reconstruction has become a natural computational model
for this setting. An unrolled network maps a finite number of iterations
of a model-based solver to a trainable architecture, interleaving learned
regularization with explicit data
consistency (DC)~\cite{schlemper2017deep,hammernik2018learning,Aggarwal_2019}.
This combination of learned image priors and encoding physics has shown
strong reconstruction performance on benchmarks such as
fastMRI~\cite{zbontar2018fastmri,sriram2020end}, making the cascade a
central unit of modern MRI reconstruction.

Most subsequent advances have refined the learned reconstruction modules
or the information exchanged between cascades. Recurrent hidden
states~\cite{yiasemis2022recurrentvariationalnetworkdeep}, recurrent
multiscale Transformers~\cite{guo2023reconformer}, hybrid multiscale
processing~\cite{fabian2022humus}, shifted-window
attention~\cite{huang2022swin}, prompting~\cite{xin2023kspacerefineimageprompting},
and structured state-space modules~\cite{Korkmaz_2025_WACV} have increased
the capacity of learned reconstruction models. Joint image and coil
sensitivity estimation~\cite{arvinte2021deepjsenseacceleratedmri,jun2021jointicnet}
and cross-cascade feature fusion~\cite{xin2024rethinking} have further
enriched the reconstruction state. These developments largely retain
the full Fourier support of the learned estimates: an early cascade can
populate the same unmeasured coefficients as the final cascade. The
intermediate reconstruction evolves with depth, while the learned
estimates remain spectrally unrestricted.

The acquisition model fixes which measurements are available, but does
not prescribe an identical support for every learned estimate. MRI
provides a natural setting in which to examine this choice. Dependencies
within local $k$-space neighborhoods~\cite{haldar2013low} and across coil
channels~\cite{shin2014calibrationless,haldar2016p} enable structured
low-rank reconstruction, while calibration methods exploit related
structure to estimate coil sensitivities~\cite{uecker2014espirit}.
The nonuniform distribution of spectral energy also motivates
variable-density sampling~\cite{lustig2007sparse}. These properties
motivate a statistical description of support expansion that accounts
for both spectral variation and dependence, rather than the number or
location of coefficients alone.

Progressive Divide-and-Conquer (PDAC)~\cite{wang2024progressive} uses a
predictor to select stage-wise spectral supports, while flow-aligned
training~\cite{qi2026unrollednetworksconditionalprobability} shapes
intermediate trajectories through cascade parameters and supervision.
For support-based progression, the central scheduling question is when
each spectral region enters the learned estimates. Early admission
retains that region over more cascades for refinement, while later
admission extends the preceding period of support-based filtering.
Distributing these transitions across a finite cascade budget therefore
determines the refinement stages available to different spectral regions.
Equal increments in window width need not add equal amounts of spectral
variation, and newly admitted coefficients can be strongly correlated
with those already present. A schedule informed by the conditional
contribution of each expansion connects this statistical structure to
the allocation of reconstruction stages.

We propose \textbf{STAIRS} (Staging the Inverse Problem via
Admissible-Support Scheduling), which constructs this allocation from
the spectral statistics of fully sampled training targets. Each cascade
applies a prescribed Fourier-support projection and conditions its
learned update on the support transition. The supports are nested and expand to the
full grid at the final cascade. Every DC step uses all acquired
measurements, while the physical sampling mask and DC formulation remain
unchanged. The complete schedule is computed before network training and
remains fixed during training and inference, without an input-dependent
support predictor.

Schedule construction uses a regularized log-determinant functional of
training-set spectral covariance. On a prescribed nested family, its
Schur-complement increments measure the contribution of newly admitted
coefficients after accounting for their correlations with the existing
support. Their normalized cumulative sum defines a monotone spectral
coordinate along the family. Stage targets allocate this coordinate
across cascades, and each target selects the smallest candidate support
that attains it. We use dyadic targets, whose successive increments
halve: early cascades admit a larger share of the coordinate, and later
cascades add progressively smaller shares while continuing to refine
the regions already admitted. We evaluate this allocation against
uniform, back-loaded, and manually designed alternatives. Because the
same targets are mapped through the coordinate of each MRI setting, one
allocation rule yields a setting-specific support path, which both host
regularizers share.

We instantiate STAIRS with Transformer-based
HUMUSNet~\cite{fabian2022humus} and an unrolled CNN with a
NAFNet-based regularizer~\cite{chen2022simple}, and evaluate single- and
multi-coil knee reconstruction on fastMRI~\cite{zbontar2018fastmri} and
Stanford2D~\cite{cheng2018stanford2dfse} with multi-coil knee MRI.
Matched-host comparisons show reconstruction gains with both hosts.
For cross-dataset evaluation, we transfer the trained reconstructor and its schedule
together, without fine-tuning or re-estimation. STAIRS improves over the
matched baselines and, under the shared HUMUSNet host, over PDAC
in both transfer directions. Under the shared HUMUSNet implementation,
STAIRS also uses fewer parameters and has lower measured latency than
PDAC.

Our contributions are as follows:
\begin{itemize}
\item We introduce STAIRS, a framework for unrolled MRI reconstruction
that couples progressive support projection along a fixed,
data-derived schedule with transition conditioning, improving the
intermediate reconstructions passed between cascades.

\item We characterize support expansion through conditional
log-determinant increments and combine the resulting spectral
coordinate with dyadic stage allocation, achieving higher
reconstruction accuracy than uniform and manually designed
schedules under a fixed backbone.

\item We show that STAIRS improves both convolutional and Transformer
backbones, retaining its reconstruction gains when the source model
and schedule are transferred unchanged across datasets. With HUMUSNet,
it also outperforms PDAC while using fewer parameters and achieving
lower inference latency.
\end{itemize}

\section{Related Work}
\label{sec:related_work}

\subsection{Learned Priors and Unrolled Reconstruction}

End-to-end unrolling embeds learned regularization and explicit data
consistency in a jointly trained reconstruction pipeline.
ADMM-Net~\cite{yang2017admmnetdeeplearningapproach},
variational networks~\cite{hammernik2018learning}, and
MoDL~\cite{Aggarwal_2019} parameterize iterations of regularized
inverse problems. Cascaded CNNs couple learned image corrections to
data consistency~\cite{schlemper2017deep}, while KIKI-Net extends
this processing across image and $k$-space domains~\cite{eo2018kiki}.
E2E-VarNet demonstrates the benefit of jointly learning sensitivity
estimation and multi-coil reconstruction~\cite{sriram2020end}.

Within this formulation, recurrent
state~\cite{yiasemis2022recurrentvariationalnetworkdeep} carries
information across iterations. Multiscale
representations~\cite{fabian2022humus} and Transformer-based
processing~\cite{guo2023reconformer,huang2022swin} enrich the learned
prior, while prompting~\cite{xin2023kspacerefineimageprompting} and
structured state-space modules~\cite{Korkmaz_2025_WACV} provide
contextual conditioning and long-range modeling, respectively.
NAFNet shows that effective restoration priors can also be built from
simple gated convolutions~\cite{chen2022simple}.
Joint image--sensitivity estimation~\cite{arvinte2021deepjsenseacceleratedmri,jun2021jointicnet}
and cross-cascade feature propagation~\cite{xin2024rethinking} further
extend the variables updated or exchanged between cascades.
STAIRS retains this end-to-end formulation: the reconstruction modules
are trained jointly along a fixed schedule, with one support projection
per cascade. Schedule construction is independent of the regularizer
architecture and complements the convolutional and Transformer hosts
considered here.

\subsection{Fourier Structure and Measurement Design}

Structured spectral models exploit linear dependencies among Fourier
coefficients to recover missing data. SAKE expresses these dependencies
through low-rank block-Hankel matrices of multi-coil
$k$-space~\cite{shin2014calibrationless}.
LORAKS relates limited image support and smooth phase to local low-rank
structure~\cite{haldar2013low}, while P-LORAKS couples this structure
across coils~\cite{haldar2016p}.
ALOHA connects transform-domain sparsity to annihilating-filter
relations~\cite{jin2016general}, and GIRAF exploits convolutional
structure to solve related low-rank recovery problems
efficiently~\cite{ongie2017fast}.
ESPIRiT uses calibration subspaces to estimate coil
sensitivities~\cite{uecker2014espirit}.
These methods use within-scan structure to constrain the reconstructed
signal or estimate its encoding model. STAIRS shares the premise of
spectral dependence, but uses covariance across training spectra to
allocate stage supports, rather than imposing a structured low-rank
constraint on each reconstruction.

Statistical criteria also guide acquisition design. Bayesian experimental
design selects sampling patterns under a probabilistic reconstruction
model~\cite{seeger2010optimization}, while OEDIPUS uses estimation
bounds under sparsity constraints~\cite{haldar2019oedipus}.
Learning-based methods optimize masks for a fixed reconstruction
rule~\cite{gozcu2018learning} or jointly optimize sampling and
reconstruction~\cite{bahadir2020deeplearningbasedoptimizationundersamplingpattern};
sequential approaches adapt acquisition to observed
data~\cite{pmlr-v158-yin21a}.
These approaches select the measurements available to a reconstructor.
STAIRS instead uses spectral statistics to allocate projection supports
after the acquisition mask has been fixed.

\subsection{Progressive Reconstruction}

PDAC~\cite{wang2024progressive} decomposes subsampling into nested reconstruction stages, combining hand-prescribed line-count budgets with input-dependent line selection and severity conditioning; it evaluates several hand-crafted budget schedules via enumeration, adopting the one that performs best empirically. 
STAIRS shares the use of nested supports and stage conditioning, but
chooses stage supports by thresholding a correlation-aware training-set
log-determinant coordinate within a prescribed family. The resulting
sizes and locations are fixed before network training and shared across
inputs and hosts.

UEPS~\cite{zhou2026uepsrobustefficientmri} also expands a central Fourier region, but couples this expansion
to increasing reconstruction resolution and incorporates acquired
measurements as the grid grows.
STAIRS keeps the reconstruction grid fixed and uses all acquired
measurements in every DC step; the schedule changes the admissible
support of each cascade, not the grid resolution.

FLAT~\cite{qi2026unrollednetworksconditionalprobability} organizes cascade trajectories through flow-derived parameters
and intermediate velocity supervision.
STAIRS instead constrains Fourier support by projection at each cascade
and retains the host reconstruction objective, without prescribing
intermediate image or velocity targets.

Coarse-to-fine reconstruction also includes learned iterates on
increasing-resolution grids~\cite{hauptmann2020multiscale}, and pyramid
recurrent reconstruction~\cite{chen2022pyramid}.
Related ideas include frequency continuation in waveform
inversion~\cite{bunks1995multiscale,sirgue2004efficient}, frequency
marching in cryo-EM~\cite{barnett2017rapid}, and the low-frequency bias
of neural networks~\cite{rahaman2019spectral}. STAIRS uses a fixed
reconstruction grid and computes the support allocation from a
correlation-aware coordinate of training spectra.

\begin{figure*}[!htbp]
    \centering
    \includegraphics[width=0.97\linewidth]{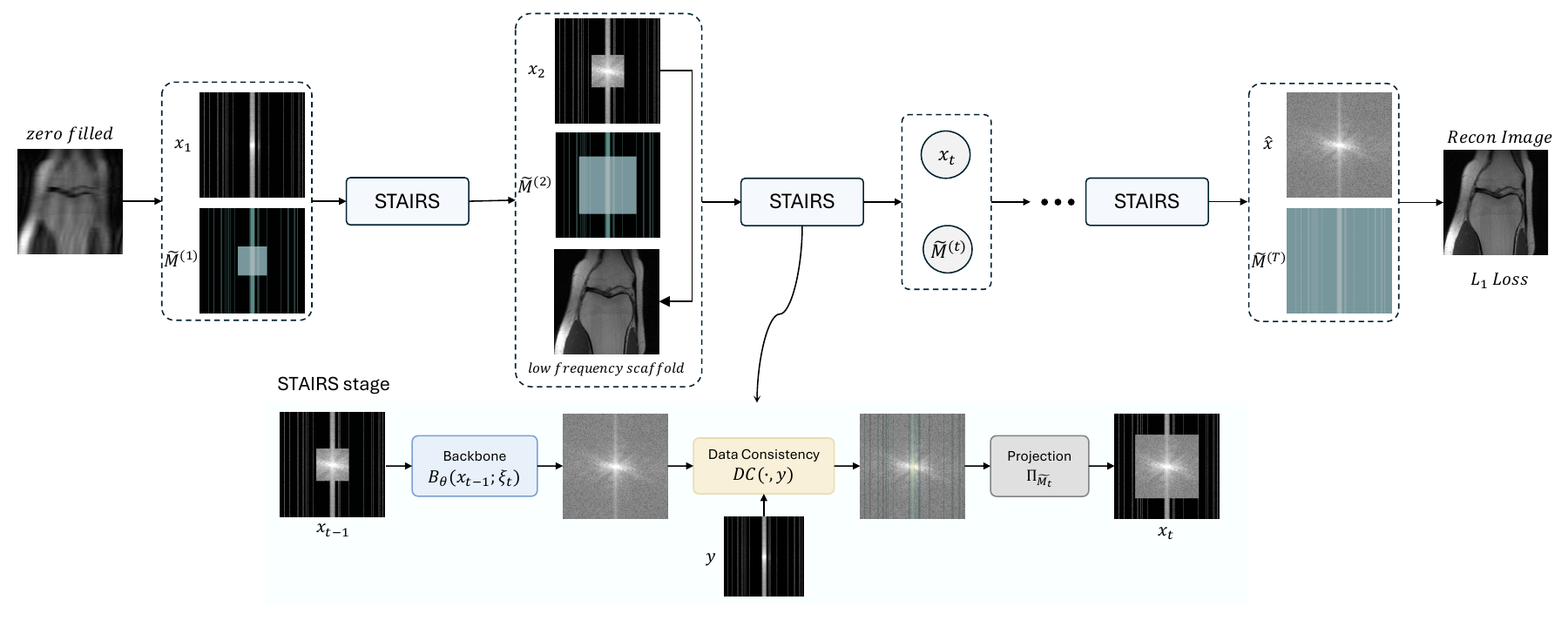}
    \caption{STAIRS overview. Each cascade applies a learned update, physical data consistency, and an admissible-support projection on a fixed reconstruction grid. The support schedule is computed once from training spectra using a correlation-aware log-determinant score and remains fixed during training and inference. All acquired measurements are used at every DC step, and the final cascade admits the full Fourier grid.}
    \label{fig:overview}
\end{figure*}

\section{Method}
\label{sec:method}

\subsection{Admissible-Support Unrolling}
\label{sec:stage_support}
Let $x\in\C^d$ denote the target image on a Cartesian Fourier grid $\Omega$. Multi-coil measurements follow
\begin{equation}
    y=P_M\F Sx+\epsilon=:A_Mx+\epsilon ,
    \label{eq:forward_model}
\end{equation}
where $S$ is the receiver-sensitivity operator, $\F$ is the Fourier transform, and $P_M$ restricts the data to the acquired index set $M$. The single-coil model is recovered by setting $S=I$. A conventional model-based reconstruction estimates $x$ from
\begin{equation}
    \min_x\; \frac{1}{2}\norm{A_Mx-y}_2^2+\lambda\Psi(x),
    \label{eq:map_objective}
\end{equation}
with $\Psi$ encoding prior information. Unrolled networks replace the optimizer of \eqref{eq:map_objective} by a finite sequence of learned updates and data-consistency operations derived from $A_M$.

STAIRS makes the Fourier support of the learned estimate depend on cascade
depth. We prescribe nested admissible supports
\begin{equation}
    K^{(1)}\subseteq K^{(2)}\subseteq\cdots\subseteq K^{(T)}=\Omega,
    \label{eq:stage_path}
\end{equation}
which specify the unmeasured coefficients admitted at each stage. At cascade $t$, the acquired coefficients and the coefficients admitted by $K^{(t)}$ form the effective support $\widetilde M^{(t)}=M\cup K^{(t)}$. Its image-domain projection is
\begin{equation}
    \Pi_t=\F^{-1}P_{\widetilde M^{(t)}}\F.
    \label{eq:projection}
\end{equation}
Writing $\xi_t=(\widetilde M^{(t-1)},\widetilde M^{(t)})$ for the support transition, the staged cascade is
\begin{equation}
    x_t=\Pi_t\!\left[
        \DC_{A_M,\mu_t}\!\left(\B_{\theta_t}(x_{t-1};\xi_t);y\right)
    \right].
    \label{eq:stage_update}
\end{equation}
The learned update $\B_{\theta_t}$ and data-consistency map
$\DC_{A_M,\mu_t}$ retain the form of the host network.
The projection $\Pi_t$ restricts the learned estimate to the
stage-admissible support after DC incorporates all acquired measurements.
Since $K^{(T)}=\Omega$, the terminal projection is the identity. Fig.~\ref{fig:overview} summarizes the reconstruction cascade and support schedule.

\subsection{Conditional Spectral Coordinate}
\label{sec:coordinate}

We select the $T$ stage supports as $K^{(t)}=K_{j_t}$ from a prescribed
family of $L+1$ nested candidates,
\begin{equation}
    K_0\subset K_1\subset\cdots\subset K_L=\Omega,
    \qquad S_j=K_j\setminus K_{j-1}.
    \label{eq:candidate_family}
\end{equation}
Here $K_0$ is the ACS window. Our default candidates are centered square
windows of increasing even width $w_j$, from $K_0$ to the full grid.
The same construction also applies to nested bands of phase-encoding
lines, which we evaluate in the schedule ablation.

All candidates use the same $N>1$ preprocessed training spectra
$k^{(n)}\in\C^d$ from fully sampled targets. Define
\begin{equation}
    u_K^{(n)}=
    \begin{bmatrix}
        \operatorname{Re}(P_Kk^{(n)})\\
        \operatorname{Im}(P_Kk^{(n)})
    \end{bmatrix}\in\R^{2|K|},
    \label{eq:real_imag_features}
\end{equation}
$\bar u_K^{(n)}=u_K^{(n)}-N^{-1}\sum_m u_K^{(m)}$, and
\begin{equation}
    \widehat\Sigma_K=\frac{1}{N-1}\sum_{n=1}^{N}
    \bar u_K^{(n)}\bar u_K^{(n)\top}.
    \label{eq:empirical_covariance}
\end{equation}
We score each candidate using a common scale $\tau>0$:
\begin{equation}
    \Phi_\tau(K)=\frac12\log\det\!\left(
        I+\tau^{-1}\widehat\Sigma_K
    \right).
    \label{eq:logdet_score}
\end{equation}

For independent $g_K\sim\mathcal N(0,\widehat\Sigma_K)$ and
$\eta_K\sim\mathcal N(0,\tau I)$, this equals the mutual information
$I(g_K;g_K+\eta_K)$: the information that a noisy observation of the
coefficients in $K$ carries about a Gaussian surrogate of the training
spectrum. Along the nested family, the increase in this score measures
the contribution of each new shell after accounting for its dependence
on the preceding support.

\begin{proposition}[Conditional spectral increment]
\label{prop:schur_increment}
For candidates evaluated from the same centered samples and scale
$\tau>0$, each shell contributes
\end{proposition}
\begin{align}
    \Delta_j&:=\Phi_\tau(K_j)-\Phi_\tau(K_{j-1})\nonumber\\
    &=\frac12\log\det\!\left(
        I+\tau^{-1}\widehat\Sigma^{(\tau)}_{S_j\mid K_{j-1}}
    \right)\ge0,
    \label{eq:schur_increment}
\end{align}
where
\begin{align}
    \widehat\Sigma^{(\tau)}_{S_j\mid K_{j-1}}
    &=\widehat\Sigma_{S_j}\nonumber\\
    &\quad-\widehat\Sigma_{S_j,K_{j-1}}
    (\tau I+\widehat\Sigma_{K_{j-1}})^{-1}
    \widehat\Sigma_{K_{j-1},S_j},
    \label{eq:regularized_schur}
\end{align}
and $\widehat\Sigma_{K,K'}$ denotes the empirical cross-covariance. 

\begin{proof}
Write $K=K_{j-1}$ and $S=S_j$, and group the coordinates of $K_j$
accordingly. Since $\tau I+\widehat\Sigma_K\succ0$, the
block-determinant identity gives
\begin{align*}
\det\!\left(I+\tau^{-1}\widehat\Sigma_{K_j}\right)
&=\det\!\left(I+\tau^{-1}\widehat\Sigma_K\right)\\
&\quad\times\det\!\left(
I+\tau^{-1}\widehat\Sigma^{(\tau)}_{S\mid K}\right).
\end{align*}
Taking half the logarithm of both sides gives the equality
in~\eqref{eq:schur_increment}. Moreover,
$\widehat\Sigma^{(\tau)}_{S\mid K}$ is the Schur complement of the
$K$-block in the positive semidefinite matrix
$\widehat\Sigma_{K_j}+\operatorname{diag}(\tau I,0)$.
Thus $\widehat\Sigma^{(\tau)}_{S\mid K}\succeq0$, implying
$\Delta_j\ge0$. Under the Gaussian model, the Gaussian conditioning
formula identifies this same matrix with
$\operatorname{Cov}(g_S\mid g_K+\eta_K)$.
\end{proof}

The conditional covariance removes the variation in $S_j$ explained by
the noisy observation of the preceding support. A shell with large total
variance can therefore contribute a small increment when its coefficients
are strongly predictable from those already included. Normalizing
the accumulated increments gives the spectral coordinate
\begin{equation}
    E_j=\frac{\Phi_\tau(K_j)-\Phi_\tau(K_0)}
    {\Phi_\tau(K_L)-\Phi_\tau(K_0)},
    \qquad j=0,\ldots,L,
    \label{eq:cumulative_coordinate}
\end{equation}
since $\Phi_\tau(K_L)>\Phi_\tau(K_0)$. By
Proposition~\ref{prop:schur_increment}, $0=E_0\le\cdots\le E_L=1$,
and $1-E_j$ is the fraction of the score beyond the reference support
that remains outside $K_j$. The remaining spectral score satisfies
$\Phi_\tau(\Omega)-\Phi_\tau(K_j)
\le\tr\widehat\Sigma_{\Omega\setminus K_j}/(2\tau)$.

\subsection{Schedule Construction}
\label{sec:scheduler}

With the candidate family parameterized by $E_j$, stage allocation
reduces to selecting cumulative targets and mapping them back to
supports. For a target $q\in(0,1]$, we use the first candidate that
reaches it:
\begin{equation}
    j^\star(q)=\min\{j\in\{1,\ldots,L\}:E_j\ge q\}.
    \label{eq:target_inversion}
\end{equation}
This rule reaches the requested spectral coordinate while admitting
only the smallest support available in the candidate family. 

Since $E_j$ is non-decreasing and the candidates are nested,
$K_{j^\star(q)}$ admits the fewest unmeasured coefficients
among the candidates with $E_j\ge q$.
By \eqref{eq:cumulative_coordinate}, its residual score satisfies
\[
    \Phi_\tau(\Omega)-\Phi_\tau(K_{j^\star(q)})
    \le (1-q)\bigl[\Phi_\tau(\Omega)-\Phi_\tau(K_0)\bigr].
\]
In the single-coil model, these unmeasured coefficients are
exactly the coordinates on which $A_M$ vanishes, so the rule
also admits the fewest unobserved directions. The inversion maps each target to a support, while the target sequence
determines when successive spectral regions enter the learned estimate
and how many later cascades are available to refine them.

We adopt a front-loaded allocation through normalized dyadic targets,
\begin{equation}
    q_t=\frac{1-2^{-t}}{1-2^{-T}},\qquad t=1,\ldots,T,
    \label{eq:dyadic_targets}
\end{equation}
and set $j_t=j^\star(q_t)$ for $t<T$, with $j_T=L$. The target
increments $q_t-q_{t-1}=2^{-t}/(1-2^{-T})$, where $q_0=0$, halve
at each cascade. This assigns larger shares of the spectral coordinate
to earlier cascades, where admitted coefficients have more subsequent
stages for refinement. Later targets add progressively smaller shares
as the support approaches the full grid. Mapping
these targets through $E_j$ determines the support sizes: front-loading
is specified in the spectral coordinate, while the resulting widths
depend on the training spectra. Discrete candidates can produce repeated indices,
and assigning $j_T=L$ ensures full terminal support even when the
score reaches a plateau earlier.

\subsection{Offline Computation and Host Integration}
\label{sec:host_integration}

To evaluate the scores without forming a covariance matrix in the
full feature dimension, we use the sample-space Gram matrix
\begin{equation}
    G_j=\frac{1}{N-1}\bar X_j\bar X_j^\top\in\R^{N\times N},
    \label{eq:sample_gram}
\end{equation}
where $\bar X_j\in\R^{N\times2|K_j|}$ contains
$\bar u_{K_j}^{(n)\top}$ as its rows. Sharing the nonzero eigenvalues
of $\widehat\Sigma_{K_j}$, this matrix gives the same score through
$\Phi_\tau(K_j)=\tfrac12\log\det(I+\tau^{-1}G_j)$.
For the retained positive eigenvalues
$0<\lambda_1\le\cdots\le\lambda_R$ of $G_L$, with $R\ge32$, we set
\begin{equation}
    \tau=\operatorname{median}\{\lambda_1,\ldots,\lambda_{m_\tau}\},
    \qquad m_\tau=\max(32,\lceil0.1R\rceil).
    \label{eq:tau_rule}
\end{equation}
Schedule sensitivity to this choice is examined in the Appendix.

\begin{algorithm}[!htbp]
\small
\caption{Offline STAIRS schedule construction}
\label{alg:stairs_scheduler}
\begin{algorithmic}[1]
\REQUIRE Training spectra $\mathcal D$; nested candidates
$\{K_j\}_{j=0}^{L}$; acquired set $M$; cascade count $T$.
\ENSURE Stage supports $\{K^{(t)}\}_{t=1}^{T}$ and effective
supports $\{\widetilde M^{(t)}\}_{t=1}^{T}$.
\STATE Center the training features and form $\bar X_L$.
\STATE Compute $G_L$ by~\eqref{eq:sample_gram} and choose
$\tau$ by~\eqref{eq:tau_rule}.
\FOR{$j=0,\ldots,L$}
    \STATE Form $G_j$ using~\eqref{eq:sample_gram}.
    \STATE $\phi_j \gets \tfrac12\log\det(I+\tau^{-1}G_j)$.
\ENDFOR
\STATE $E_j \gets (\phi_j-\phi_0)/(\phi_L-\phi_0)$,
$j=0,\ldots,L$.
\FOR{$t=1,\ldots,T-1$}
    \STATE $q_t \gets (1-2^{-t})/(1-2^{-T})$.
    \STATE $j_t \gets \min\{j\in\{1,\ldots,L\}:E_j\ge q_t\}$.
    \STATE $K^{(t)} \gets K_{j_t}$.
\ENDFOR
\STATE $K^{(T)} \gets K_L=\Omega$.
\STATE $\widetilde M^{(t)} \gets M\cup K^{(t)}$,
$t=1,\ldots,T$.
\end{algorithmic}
\end{algorithm}

Algorithm~\ref{alg:stairs_scheduler} summarizes the offline schedule construction.
The schedule is computed once from the training partition of each
dataset and acquisition setting, shared across both hosts, and fixed
during network training and inference. For cross-dataset evaluation, we retain the schedule from the
source dataset.

In full STAIRS, transition conditioning makes the support change
explicit to the learned update. The descriptor $\xi_t$ contains the
preceding and current supports, identifying the region carried forward
and the coefficients newly admitted. A mask encoder embeds $\xi_t$
for feature-wise affine modulation~\cite{perez2017filmvisualreasoninggeneral},
applied to Transformer features in HUMUSNet~\cite{fabian2022humus}
and normalized convolutional features in the NAFNet-based unrolled
host~\cite{chen2022simple}. The cascades are trained jointly using the
host reconstruction objective along the precomputed schedule. The
support-only control removes conditioning while retaining the
projections; the conditioning-only control uses the same schedule
descriptors with $\Pi_t=I$ at every cascade.

\section{Experimental Protocol}
\label{sec:experiments}

\subsection{Data and Acquisition}
\label{sec:data}
We evaluate on the fastMRI knee benchmark~\cite{zbontar2018fastmri} and Stanford2D FSE~\cite{cheng2018stanford2dfse}. We use the 973 training and 199 validation volumes of fastMRI. Single-coil images are center-cropped to $320\times320$ and reconstructed at $4\times$ and $8\times$ acceleration, using central fractions of $8\%$ and $4\%$, respectively. Multi-coil images are cropped to $384\times384$ and evaluated at $8\times$ acceleration with a $4\%$ central fraction.

Stanford2D contains 89 fully sampled volumes. We follow the fixed volume-level 80/20 split used by HUMUSNet~\cite{fabian2022humus} and PDAC~\cite{wang2024progressive}, crop images to $384\times384$, and use the same $8\times$/4\% acquisition setting. For each dataset and acceleration factor, the support schedule is computed exclusively from the corresponding training partition.

FastMRI+~\cite{zhao2022fastmri+} adds slice-level bounding boxes, drawn by a radiologist around pathology, to the fastMRI knee volumes. We use them to test whether the gains reach clinically relevant regions.

\subsection{Hosts and Optimization}
\label{sec:hosts}
STAIRS is integrated with two unrolled hosts whose learned regularizers have different architectures: Transformer-based HUMUSNet~\cite{fabian2022humus} and convolutional NAFNet~\cite{chen2022simple}. NAFNet supplies the convolutional learned update within the unrolled MRI network. For multi-coil data, both hosts adopt the sensitivity estimation and $k$-space data consistency of E2E-VarNet~\cite{sriram2020end}. Each uses eight cascades, an image-domain $\ell_1$ objective, and AdamW. HUMUSNet is trained for 40 epochs in the single-coil experiments and 50 epochs in the multi-coil experiments; its learning rate starts at $10^{-4}$ and is reduced to $10^{-5}$ over the final five epochs. NAFNet is trained for 20 and 50 epochs, respectively, with cosine decay from $2\times10^{-4}$ to $10^{-5}$. Batch sizes are four for fastMRI and one for Stanford2D. All runs use an NVIDIA H200 GPU.

Within each host, the baseline and STAIRS share the cascade count, data split, sampling masks, initialization, loss, optimizer, training schedule, and evaluation code. STAIRS adds only the scheduled support projections and transition conditioning described in Sec.~\ref{sec:host_integration}.

\begin{figure}[!htbp]
  \centering
  \includegraphics[width=\columnwidth]{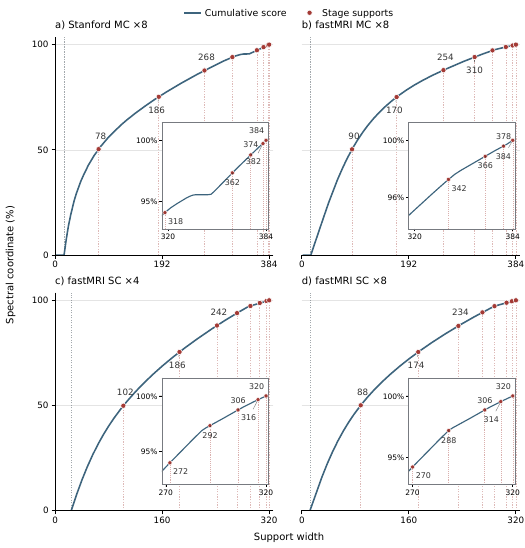}
  \caption{Empirical spectral coordinates and realized support schedules for four
MRI settings. Curves show the spectral coordinate $E_j$ as a function
of support width; markers indicate the selected stage supports. Insets
magnify the region near full support.}
  \label{fig:realized-schedules}
\end{figure}

\section{Results}

\subsection{Realized Support Schedules}
\label{sec:realized-schedules}

Figure~\ref{fig:realized-schedules} shows how a shared set of
dyadic targets is realized as distinct support paths. On the
$384\times384$ multi-coil grids, the first target
($q_1\simeq0.502$) selects widths of 78 for Stanford2D and 90
for fastMRI. At the second cascade, the widths are 186 and 170,
respectively. The reversal in ordering shows how the training-set
spectral profiles shape support expansion across stages.

Across all four settings, the later supports cluster near the
full grid as the increments in the spectral coordinate become
smaller. The resulting schedules combine large early advances
in the coordinate with finer support expansion in the later
cascades, reaching full support at C8.

\subsection{Effect of Support Scheduling}
\label{sec:projection-effect}

\begin{table}[!t]
\centering
\caption{Quantitative results of $8\times$ accelerated multi-coil MRI reconstruction on fastMRI knee and Stanford2D. Best results within each host are in \best{teal}.}
\label{tab:results}
\footnotesize
\setlength{\tabcolsep}{3pt}
\resizebox{\columnwidth}{!}{%
\begin{tabular}{@{}lcccccc@{}}
\toprule
\multirow{2}{*}{Method}
  & \multicolumn{3}{c}{fastMRI knee} & \multicolumn{3}{c}{Stanford2D} \\
\cmidrule(lr){2-4} \cmidrule(lr){5-7}
  & PSNR$\uparrow$ & SSIM$\uparrow$ & NMSE$\downarrow$
  & PSNR$\uparrow$ & SSIM$\uparrow$ & NMSE$\downarrow$ \\
\midrule
Zero-filled                      & 27.42 & 0.7000 & 0.0704 & 29.24 & 0.7175 & 0.0948 \\
U-Net~\cite{ronneberger2015u}    & 34.18 & 0.8601 & 0.0151 & 33.77 & 0.8953 & 0.0333 \\
E2E-VarNet~\cite{sriram2020end}  & 36.81 & 0.8873 & 0.0090 & 35.80 & 0.9118 & 0.0269 \\
\midrule
HUMUSNet~\cite{fabian2022humus}  & 36.84 & 0.8879 & 0.0092 & 37.36 & 0.9266 & 0.0183 \\
\quad + PDAC~\cite{wang2024progressive}
                                 & 37.12 & 0.8905 & 0.0085 & 37.77 & 0.9269 & 0.0169 \\
\quad + \textbf{STAIRS}          & \best{37.28} & \best{0.8925} & \best{0.0082} & \best{37.96} & \best{0.9316} & \best{0.0163} \\
\midrule
NAFNet~\cite{chen2022simple}     & 37.13 & 0.8911 & 0.0085 & 37.10 & 0.9261 & 0.0198 \\
\quad + \textbf{STAIRS}          & \best{37.33} & \best{0.8930} & \best{0.0081} & \best{37.22} & \best{0.9272} & \best{0.0190} \\
\bottomrule
\end{tabular}%
}
\end{table}

\begin{table*}[!b]
\centering
\caption{ROI-based results of $8\times$ multi-coil reconstruction within fastMRI+ pathology bounding boxes, pooled and per anatomical group. Best results within each host are in \best{teal}.}
\label{tab:fastmri_plus_roi}
\footnotesize
\setlength{\tabcolsep}{3pt}
\resizebox{\textwidth}{!}{%
\begin{tabular}{@{}l*{15}{c}@{}}
\toprule
\multirow{3}{*}{Method}
  & \multicolumn{3}{c}{All ROIs}
  & \multicolumn{3}{c}{Meniscus \& cartilage}
  & \multicolumn{3}{c}{Ligament \& support}
  & \multicolumn{3}{c}{Bone}
  & \multicolumn{3}{c}{Joint \& soft tissue} \\
  & \multicolumn{3}{c}{($n{=}2498$)}
  & \multicolumn{3}{c}{($n{=}1414$)}
  & \multicolumn{3}{c}{($n{=}347$)}
  & \multicolumn{3}{c}{($n{=}360$)}
  & \multicolumn{3}{c}{($n{=}377$)} \\
\cmidrule(lr){2-4} \cmidrule(lr){5-7} \cmidrule(lr){8-10} \cmidrule(lr){11-13} \cmidrule(lr){14-16}
  & PSNR$\uparrow$ & SSIM$\uparrow$ & NMSE$\downarrow$
  & PSNR$\uparrow$ & SSIM$\uparrow$ & NMSE$\downarrow$
  & PSNR$\uparrow$ & SSIM$\uparrow$ & NMSE$\downarrow$
  & PSNR$\uparrow$ & SSIM$\uparrow$ & NMSE$\downarrow$
  & PSNR$\uparrow$ & SSIM$\uparrow$ & NMSE$\downarrow$ \\
\midrule
HUMUSNet
  & 35.39 & 0.8922 & 0.0184
  & 35.89 & 0.9131 & 0.0206
  & 35.04 & 0.8739 & 0.0207
  & 34.36 & 0.8196 & 0.0205
  & 34.84 & 0.9003 & 0.0064 \\
\quad + PDAC
  & 35.44 & 0.8911 & 0.0183
  & 35.92 & 0.9121 & 0.0205
  & 35.11 & 0.8733 & 0.0202
  & 34.40 & 0.8162 & 0.0204
  & 34.90 & 0.9001 & 0.0064 \\
\quad + \textbf{STAIRS}
  & \best{35.64} & \best{0.8950} & \best{0.0174}
  & \best{36.12} & \best{0.9154} & \best{0.0196}
  & \best{35.33} & \best{0.8785} & \best{0.0191}
  & \best{34.58} & \best{0.8218} & \best{0.0194}
  & \best{35.11} & \best{0.9034} & \best{0.0061} \\
\midrule
NAFNet
  & 35.51 & 0.8943 & 0.0181
  & 35.98 & 0.9143 & 0.0203
  & 35.16 & 0.8787 & 0.0194
  & 34.52 & \best{0.8239} & 0.0198
  & 35.02 & 0.9029 & 0.0062 \\
\quad + \textbf{STAIRS}
  & \best{35.67} & \best{0.8957} & \best{0.0173}
  & \best{36.15} & \best{0.9160} & \best{0.0194}
  & \best{35.45} & \best{0.8800} & \best{0.0186}
  & \best{34.56} & 0.8222 & \best{0.0196}
  & \best{35.18} & \best{0.9043} & \best{0.0060} \\
\bottomrule
\end{tabular}%
}
\end{table*}

\begin{figure*}[!t]
  \centering
  \includegraphics[width=\textwidth]{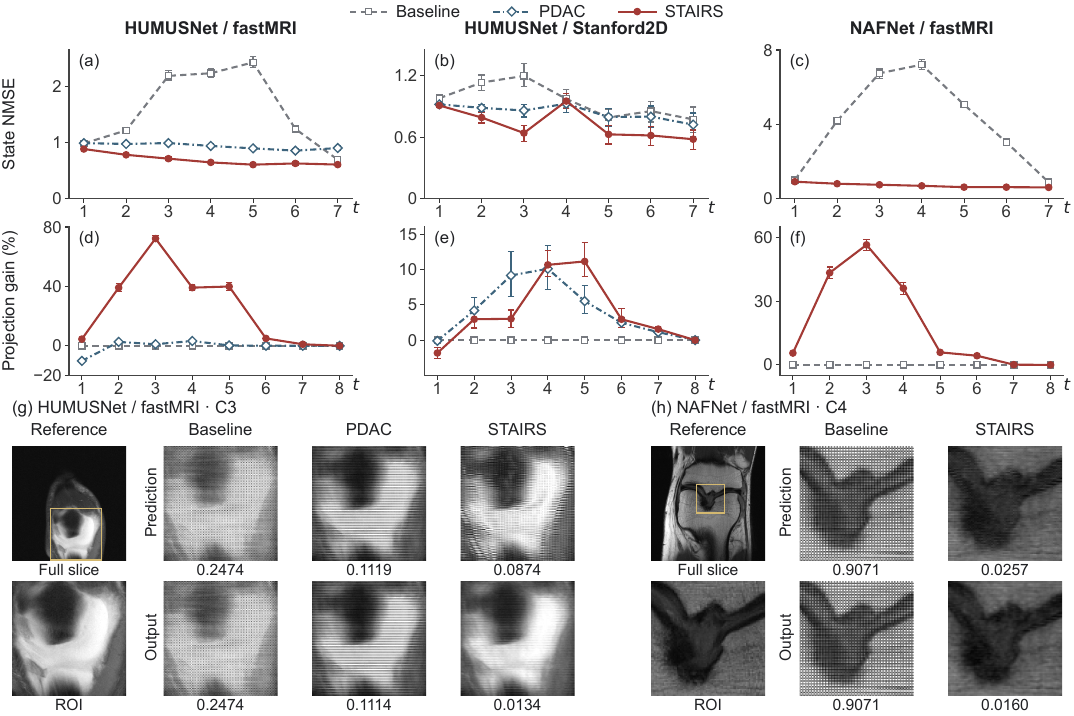}
\caption{Stage-wise reconstruction dynamics in $8\times$ multi-coil MRI.
(a)--(c) Intermediate $k$-space NMSE and (d)--(f) relative NMSE reduction
after projection; error bars show pointwise 95\% volume-bootstrap
confidence intervals. (g), (h) Projection effects in effusion and ACL
ROIs, with image-space NMSE reported below each reconstructed crop.}
\label{fig:projection-process}
\end{figure*}

STAIRS maintains more accurate intermediate reconstructions than the
baseline networks throughout support expansion
(Fig.~\ref{fig:projection-process}(a)--(c)). We
evaluate the states passed to the next cascade using NMSE over all
unmeasured complex coil $k$-space coefficients. Across all three settings,
STAIRS achieves lower mean intermediate NMSE than the baseline
at every constrained stage C1--C7. With HUMUSNet, it also outperforms
PDAC throughout C1--C7 on fastMRI and at six of the seven stages on
Stanford2D.

The projection gains reveal repeated error removal along these
trajectories (Fig.~\ref{fig:projection-process}(d)--(f)). Each gain is
the relative NMSE reduction between the same prediction before and
after projection. On fastMRI, STAIRS yields positive corrections
throughout C1--C7 with both hosts. The corrections are strongest in
the middle cascades, where the corresponding baselines develop broad
error peaks. With HUMUSNet, STAIRS removes approximately 40\%--70\%
of the prediction error across C2--C5, and its projection gain exceeds
PDAC's at every constrained stage. PDAC provides much smaller
corrections and increases error at C1.

As the schedule admits more of the unmeasured spectrum in later
cascades, the projection gains decrease while the intermediate states
remain more accurate than those of the baseline networks. On Stanford2D,
STAIRS exceeds PDAC's projection gain throughout C4--C7, while its mean
intermediate NMSE is approximately 20\%--23\% lower than PDAC's at
C5--C7 (Fig.~\ref{fig:projection-process}(b) and (e)). At C8, full
support makes projection the identity and its gain is zero.

These corrections are also visible in the fastMRI ROI examples. In the
C3 effusion ROI with HUMUSNet, the STAIRS projection suppresses stripe-like
artifacts and brings the curved fluid contour closer to the reference
(Fig.~\ref{fig:projection-process}(g)). Image-space NMSE within this ROI
falls by 84.6\% after projection, compared with 0.45\% for PDAC. With
NAFNet, the C4 projection reduces residual structured artifacts around
the ACL and lowers ROI NMSE by 37.9\%
(Fig.~\ref{fig:projection-process}(h)).

\subsection{Multi-coil Reconstruction}

\begin{figure*}[t]
\centering
\includegraphics[width=\linewidth]{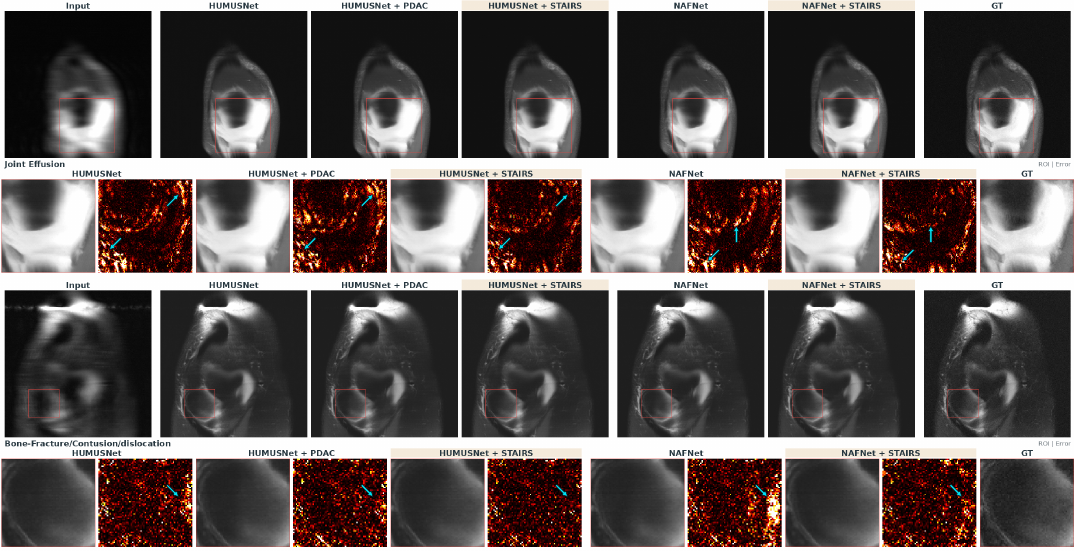}
\caption{Qualitative comparison of $8\times$ multi-coil reconstructions on fastMRI
knee. Radiologist-annotated ROIs from fastMRI+ are magnified alongside their
error maps. The top ROI shows the curved border of a joint effusion; the
bottom ROI shows bone contours in a bone-injury case. Cyan arrows highlight local error reductions with STAIRS.}
\label{fig:qual_mc}
\end{figure*}

STAIRS improves both hosts on both datasets in all three metrics, with PSNR
gains of 0.12--0.60~dB (Table~\ref{tab:results}). A host with STAIRS gives
the best value in every column, and, under the shared HUMUSNet host, STAIRS
also outperforms PDAC throughout. The gains with both Transformer and convolutional regularizers show that
support allocation complements these two forms of learned regularization. 

The gains extend to radiologist-annotated pathology
(Table~\ref{tab:fastmri_plus_roi}). Pooled over all ROIs, STAIRS improves
both hosts in all three metrics, and NAFNet+STAIRS is the best model; under
HUMUSNet, the improvement holds in every anatomical group. PDAC behaves
differently: its 0.28~dB image-level gain over HUMUSNet shrinks to 0.05~dB
within the ROIs, and its SSIM falls below that of the host in every group,
whereas STAIRS retains a 0.25~dB gain. Fig.~\ref{fig:qual_mc} shows representative reconstructions within fastMRI+ pathology ROIs, where STAIRS recovers the annotated structures with lower error than the baseline networks.

\begin{figure*}[t]
  \centering
  \includegraphics[width=\textwidth]{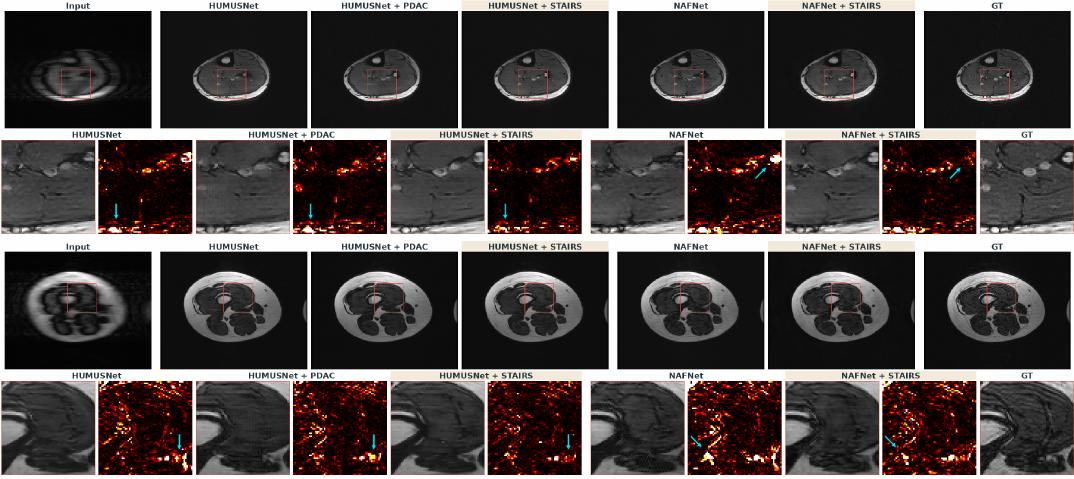}
  \caption{Qualitative comparison of $8\times$ multi-coil reconstructions on
Stanford2D. The top ROI shows small high-signal structures and tissue
borders; the bottom ROI shows curved tissue interfaces. Cyan arrows highlight local error reductions with STAIRS.}
\label{fig:qual_stanford}
\end{figure*}

On Stanford2D (Fig.~\ref{fig:qual_stanford}), adding STAIRS to HUMUSNet
improves the delineation of small bright structures in the upper example
and reduces residual errors along the curved tissue boundaries in the
lower example.

For multi-coil reconstruction with HUMUSNet, STAIRS uses 119.08M
parameters, 15.9\% fewer than PDAC's 141.60M. With fixed inputs
resident on the GPU, its model-only forward pass takes 185.79~ms,
compared with 194.64~ms for PDAC and 177.94~ms for the baseline. At inference, STAIRS uses the precomputed support schedule directly,
eliminating the auxiliary support-prediction network.

\subsection{Cross-Dataset Transfer}
\label{sec:cross-dataset-transfer}

For cross-dataset evaluation between fastMRI and Stanford2D, we transfer the trained reconstructor and its precomputed schedule together. STAIRS improves both
hosts in every metric and direction, with PSNR gains of 0.15--1.06~dB,
and HUMUSNet+STAIRS achieves the best results throughout
(Table~\ref{tab:cross_domain_mc}). Under the shared HUMUSNet host,
STAIRS exceeds PDAC by 1.36~dB for fastMRI-to-Stanford2D transfer and
1.52~dB in the reverse direction. PDAC improves HUMUSNet in-domain
but falls below the baseline in both transfer directions.
STAIRS thus retains its advantage over each host when the support
schedule computed from source training spectra is used on the
target dataset.

\begin{table}[!htbp]
\centering
\caption{Cross-dataset evaluation of $8\times$ multi-coil reconstruction (A$\rightarrow$B: trained on A, tested on B). Best results within each host are in \best{teal}.}
\label{tab:cross_domain_mc}
\footnotesize
\setlength{\tabcolsep}{3pt}
\resizebox{\columnwidth}{!}{%
\begin{tabular}{@{}lcccccc@{}}
\toprule
\multirow{2}{*}{Method}
  & \multicolumn{3}{c}{fastMRI$\,\rightarrow\,$Stanford2D}
  & \multicolumn{3}{c}{Stanford2D$\,\rightarrow\,$fastMRI} \\
\cmidrule(lr){2-4} \cmidrule(lr){5-7}
  & PSNR$\uparrow$ & SSIM$\uparrow$ & NMSE$\downarrow$
  & PSNR$\uparrow$ & SSIM$\uparrow$ & NMSE$\downarrow$ \\
\midrule
HUMUSNet                & 34.14 & 0.8703 & 0.0382 & 33.62 & 0.8516 & 0.0171 \\
\quad + PDAC            & 33.57 & 0.8674 & 0.0443 & 32.25 & 0.8227 & 0.0232 \\
\quad + \textbf{STAIRS} & \best{34.93} & \best{0.8901} & \best{0.0343} & \best{33.77} & \best{0.8557} & \best{0.0167} \\
\midrule
NAFNet                  & 33.77 & 0.8830 & 0.0440 & 32.16 & 0.8287 & 0.0236 \\
\quad + \textbf{STAIRS} & \best{34.50} & \best{0.8899} & \best{0.0368} & \best{33.22} & \best{0.8468} & \best{0.0188} \\
\bottomrule
\end{tabular}%
}
\end{table}

\subsection{Single-Coil Reconstruction}

\begin{figure*}[t]
  \centering
  \includegraphics[width=\textwidth]{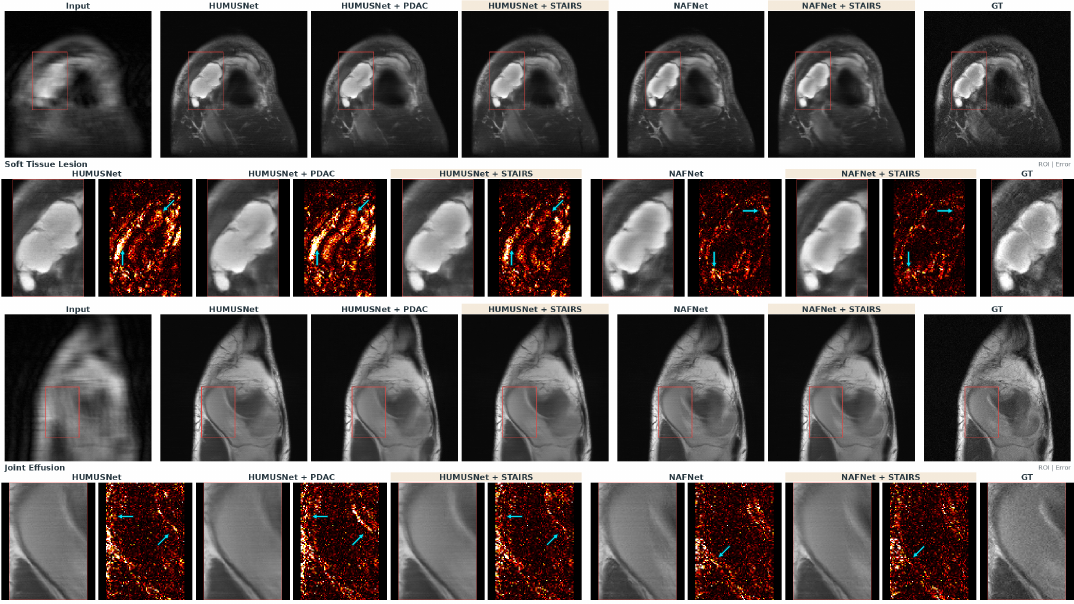}
  \caption{Qualitative comparison of $8\times$ single-coil reconstructions on fastMRI
knee. The top ROI shows the lobulated outline of a soft-tissue lesion; the
bottom ROI shows narrow tissue interfaces in a joint-effusion case. Cyan arrows highlight local error reductions with STAIRS.}
\label{fig:qual_sc}
\end{figure*}

Single-coil reconstruction sets $S=I$ in \eqref{eq:forward_model}, which
removes the coil sensitivities and their estimation. DC then modifies only
the measured $k$-space coefficients, so every unmeasured coefficient is
recovered by the learned updates alone, from the cascade whose support
first admits it. We therefore use this setting to test the schedule on the
unrolled inverse problem itself. STAIRS improves both hosts at $4\times$
and $8\times$, and NAFNet+STAIRS achieves the best results in every column
(Table~\ref{tab:recon_single}). Representative $8\times$ reconstructions
are shown in Fig.~\ref{fig:qual_sc}. The gain of the schedule thus does not rely
on coil sensitivities.

\begin{table}[!htbp]
\centering
\caption{Quantitative results of $4\times$ and $8\times$ accelerated single-coil MRI reconstruction on fastMRI knee. Best results within each host are in \best{teal}.}
\label{tab:recon_single}
\footnotesize
\setlength{\tabcolsep}{3pt}
\resizebox{\columnwidth}{!}{%
\begin{tabular}{@{}lcccccc@{}}
\toprule
\multirow{2}{*}{Method}
  & \multicolumn{3}{c}{$4\times$} & \multicolumn{3}{c}{$8\times$} \\
\cmidrule(lr){2-4} \cmidrule(lr){5-7}
  & PSNR$\uparrow$ & SSIM$\uparrow$ & NMSE$\downarrow$
  & PSNR$\uparrow$ & SSIM$\uparrow$ & NMSE$\downarrow$ \\
\midrule
Zero-filled                             & 29.49 & 0.6541 & 0.0532 & 26.84 & 0.5500 & 0.0893 \\
U-Net~\cite{ronneberger2015u}           & 31.88 & 0.7142 & 0.0357 & 29.78 & 0.6424 & 0.0511 \\
D5C5~\cite{schlemper2017deep}           & 32.25 & 0.7256 & 0.0332 & 29.65 & 0.6457 & 0.0512 \\
OUCR~\cite{guo2021over}                 & 32.61 & 0.7354 & 0.0315 & 30.59 & 0.6634 & 0.0443 \\
ReconFormer~\cite{guo2023reconformer}   & 32.73 & 0.7383 & 0.0310 & 30.89 & 0.6697 & 0.0429 \\
FMT-Net~\cite{yi2023frequency}          & 32.55 & 0.7323 & 0.0423 & 30.99 & 0.6656 & 0.0536 \\
FPS-Former~\cite{meng2025boosting}      & 32.51 & 0.7337 & 0.0316 & 31.03 & 0.6692 & 0.0408 \\
\midrule
HUMUSNet~\cite{fabian2022humus}         & 32.37 & 0.7221 & 0.0323 & 31.04 & 0.6722 & 0.0410 \\
\quad + PDAC~\cite{wang2024progressive} & 32.73 & 0.7376 & 0.0310 & 31.16 & 0.6739 & 0.0407 \\
\quad + \textbf{STAIRS}                 & \best{32.76} & \best{0.7394} & \best{0.0308} & \best{31.24} & \best{0.6786} & \best{0.0401} \\
\midrule
NAFNet~\cite{chen2022simple}            & 32.74 & 0.7385 & 0.0309 & 31.22 & 0.6767 & 0.0402 \\
\quad + \textbf{STAIRS}                 & \best{32.82} & \best{0.7411} & \best{0.0306} & \best{31.32} & \best{0.6799} & \best{0.0396} \\
\bottomrule
\end{tabular}%
}
\end{table}

\subsection{Ablation Studies}
\label{sec:ablation}

\begin{table}[!b]
\centering
\caption{Ablation of support projection and transition conditioning (Sec.~\ref{sec:host_integration}) for $8\times$ multi-coil reconstruction on Stanford2D with the HUMUSNet host. Best results are in \best{teal}.}
\label{tab:stage_support_factorized_ablation}
\footnotesize
\setlength{\tabcolsep}{3pt}
\begin{tabular*}{\columnwidth}{@{\extracolsep{\fill}}lccccc@{}}
\toprule
Variant & Support & Conditioning & PSNR$\uparrow$ & SSIM$\uparrow$ & NMSE$\downarrow$ \\
\midrule
HUMUSNet          &            &            & 37.36 & 0.9266 & 0.0183 \\
Conditioning only &            & \checkmark & 37.26 & 0.9144 & 0.0189 \\
Support only      & \checkmark &            & 37.55 & 0.9282 & 0.0176 \\
\textbf{STAIRS}   & \checkmark & \checkmark & \best{37.96} & \best{0.9316} & \best{0.0163} \\
\bottomrule
\end{tabular*}
\end{table}

Scheduled support and transition conditioning play complementary roles
(Table~\ref{tab:stage_support_factorized_ablation}). Support projection
alone raises PSNR from 37.36 to 37.55~dB and improves SSIM and NMSE,
without adding parameters. Conditioning the updates on these enforced
support transitions brings a further 0.41~dB gain, reaching 37.96~dB
with the best SSIM and NMSE. The conditioning-only control has the
same parameter count as full STAIRS but retains full support at every
cascade; it achieves 37.26~dB and underperforms the host on all three
metrics. The schedule changes the spectral task presented to each
stage, and conditioning helps the learned update respond to that
change.

\begin{figure}[!b]
  \centering
  \includegraphics[width=\columnwidth]{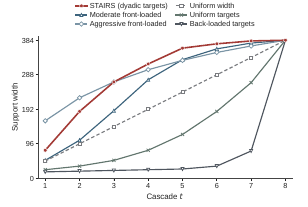}
  \caption{Expansion of square supports under alternative allocations on Stanford2D MC$\times$8. Solid curves use spectral targets; the dashed uniform-width curve is a geometric reference.}
  \label{fig:schedule_widths}
\end{figure}

\begin{table}[!htbp]
\centering
\caption{Ablation of stage allocation with square and line-band supports
for $8\times$ multi-coil reconstruction on Stanford2D with the
HUMUSNet host. Best results within each support family are
highlighted in \best{teal}.}
\label{tab:ablation_schedule_stanford}
\footnotesize
\begin{tabular*}{\columnwidth}{@{\extracolsep{\fill}}lccc@{}}
\toprule
Schedule & PSNR$\uparrow$ & SSIM$\uparrow$ & NMSE$\downarrow$ \\
\midrule
Baseline (full support) & 37.36 & 0.9266 & 0.0183 \\
\midrule
\multicolumn{4}{@{}l}{\textit{Square windows}}\\
\addlinespace[1pt]
Uniform width & 37.23 & 0.9226 & 0.0192 \\
Back-loaded targets & 36.41 & 0.9196 & 0.0227 \\
Uniform targets & 36.96 & 0.9229 & 0.0204 \\
Moderate front-loaded & 37.51 & 0.9275 & 0.0181 \\
Aggressive front-loaded & 37.53 & 0.9261 & 0.0182 \\
\textbf{STAIRS} (dyadic targets) & \best{37.96} & \best{0.9316} & \best{0.0163} \\
\midrule
\multicolumn{4}{@{}l}{\textit{Line bands}}\\
\addlinespace[1pt]
Uniform width & 36.41 & 0.9137 & 0.0226 \\
Uniform targets & 36.06 & 0.9125 & 0.0249 \\
\textbf{STAIRS} (dyadic targets) & \best{37.72} & \best{0.9294} & \best{0.0172} \\
\bottomrule
\end{tabular*}
\end{table}

We next compare how support expansion is allocated across cascades
(Fig.~\ref{fig:schedule_widths}, Table~\ref{tab:ablation_schedule_stanford}).
Within the square-window family, all target-based variants use the same
spectral mapping, with uniform width as a geometric reference. Uniform width,
uniform targets, and back-loaded targets all underperform the baseline
on every metric, whereas both manual front-loaded allocations improve PSNR and
NMSE over it. The comparison therefore favors admitting a larger share of the
spectral coordinate early. How expansion is distributed within a front-loaded
schedule also matters: relative to the aggressive front-loaded profile, STAIRS
uses narrower supports at C1--C2, the same support at C3, and wider supports
at C4--C7, and gains 0.43~dB PSNR. Its dyadic targets thus restrict the
support more strongly than this profile at the start and refine a broader
spectral region at C4--C7, yielding the best results on all three metrics
among the square-window schedules.

We also evaluate the allocation on nested bands of phase-encoding
lines (Table~\ref{tab:ablation_schedule_stanford}). With line-band
supports, STAIRS reaches 37.72\,dB PSNR, exceeding uniform-width
and uniform-target schedules by 1.31 and 1.66\,dB, respectively,
with corresponding improvements in SSIM and NMSE. Both support
families thus favor dyadic allocation over uniform expansion,
with square windows yielding the best overall reconstruction.

\section{Discussion}
\label{sec:discussion}

STAIRS makes the allocation of Fourier support across cascades a design
variable in unrolled MRI reconstruction. The learned regularizer
determines how an estimate is updated, and the support schedule determines
which reconstructed spectral components are retained after data
consistency. The gains with HUMUSNet and the NAFNet-based host in both
single- and multi-coil reconstruction show that support allocation
complements different regularizer architectures
(Tables~\ref{tab:results} and~\ref{tab:recon_single}). The reconstruction
path can thus be designed alongside the update that operates on it.

Support scheduling acts on the errors propagated between cascades.
For HUMUSNet on fastMRI MC$\times$8, the scheduled projection reduces
the error of the same network prediction at every constrained stage,
and STAIRS maintains lower intermediate errors than the baseline
and PDAC~\cite{wang2024progressive}
(Fig.~\ref{fig:projection-process}). The benefit is visible in both the
local effect of projection and the quality of the resulting intermediate
estimates. In the ablation study, scheduled support improves
reconstruction on its own, while transition conditioning strengthens the
gain when paired with the scheduled projections
(Table~\ref{tab:stage_support_factorized_ablation}). The support
transition gives conditioning a concrete change in the reconstruction
task to encode.

The schedule allocates refinement time to the unmeasured spectrum.
Early admission allows reconstructed spectral content to remain in the
projected estimates over more cascades, providing additional stages of
refinement. Delayed admission prolongs the filtering imposed by the
support restriction. The schedule ablations favor dyadic allocation
over uniform schedules for both square and line-band supports, and
over aggressive front-loading for square windows
(Table~\ref{tab:ablation_schedule_stanford}). Within each support
family, the target-based variants share the same spectral coordinate,
so their differences reveal the importance of allocation across
depth. The conditional spectral coordinate converts this common
allocation rule into support widths that follow the training
spectra of each setting (Fig.~\ref{fig:realized-schedules}).

The cross-dataset results give the fixed schedule a practical role
beyond in-domain reconstruction. STAIRS improves both hosts in both
directions between fastMRI and Stanford2D with the source reconstructor
and schedule transferred together (Table~\ref{tab:cross_domain_mc}).
Under the shared HUMUSNet host, it also exceeds PDAC. A schedule derived from training spectral statistics can therefore
remain effective across datasets, while the learned updates respond
to the measurements of each scan. Both hosts use this precomputed
schedule without an auxiliary support-prediction network.

The pooled fastMRI+ gains show that the benefit extends to annotated
pathology regions (Table~\ref{tab:fastmri_plus_roi}), connecting support
scheduling to fidelity at both image and regional scales. A natural next
step is to optimize stage targets jointly with cascade depth, using the
conditional spectral coordinate to distribute refinement across the
spectrum. Whole-image and ROI evaluations provide complementary criteria
for selecting those allocations.


\appendix
\subsection*{Sensitivity of the Estimated Schedule to \texorpdfstring{$\tau$}{tau}}

The spectral coordinate yields consistent support schedules across
alternative lower-tail estimates of $\tau$
(Table~\ref{tab:tau_sensitivity_stanford}).
Replacing the median in~\eqref{eq:tau_rule} with the mean leaves every
stage support unchanged, while the 25th and 75th percentiles change
the selected widths by at most four pixels.

\begin{table}[H]
\centering
\caption{Sensitivity of the Stanford2D multi-coil schedule to the
lower-tail estimate of $\tau$. Stage widths and their maximum absolute
changes from the default median schedule are in pixels.}
\label{tab:tau_sensitivity_stanford}
\small
\setlength{\tabcolsep}{3pt}
\renewcommand{\arraystretch}{1.08}
\begin{tabular*}{\columnwidth}{@{\extracolsep{\fill}}lrrrr@{}}
\toprule
 & Median & Mean & 25th pct. & 75th pct. \\
\midrule
$\tau\;(\times10^{-5})$ & 4.825 & 4.668 & 3.707 & 5.695 \\
\midrule
C1 & 78  & 78  & 76  & 78  \\
C2 & 186 & 186 & 182 & 190 \\
C3 & 268 & 268 & 266 & 270 \\
C4 & 318 & 318 & 316 & 320 \\
C5 & 362 & 362 & 360 & 362 \\
C6 & 374 & 374 & 374 & 374 \\
C7 & 382 & 382 & 380 & 382 \\
C8 & 384 & 384 & 384 & 384 \\
\midrule
$\max_t|\Delta w_t|$ & 0 & 0 & 4 & 4 \\
\bottomrule
\end{tabular*}
\end{table}

\bibliographystyle{IEEEtran}
\bibliography{ref}

%

\end{document}